\documentclass{article}
\usepackage{ijcai23}

\usepackage{times}
\usepackage{soul}
\usepackage{url}
\usepackage[hidelinks]{hyperref}
\usepackage[utf8]{inputenc}
\usepackage[small]{caption}
\usepackage{graphicx}
\usepackage{amsmath}
\usepackage{amsthm}
\usepackage{booktabs}
\usepackage{algorithm}
\usepackage{algorithmic}
\usepackage[switch]{lineno}

\usepackage{multirow}
\usepackage{epsfig}
\usepackage{graphics}
\usepackage{pifont}
\usepackage{amssymb}
\usepackage{newfloat}
\usepackage{listings}
\usepackage{color}

\title{TPS++: Attention-Enhanced Thin-Plate Spline for Scene Text Recognition}

\author{
Tianlun Zheng$^1$
\and
Zhineng Chen$^1$\footnote{Corresponding Author}\and
Jinfeng Bai$^{2}$\and
Hongtao Xie$^3$\and
Yu-Gang Jiang$^1$
\affiliations
$^1$Shanghai Collaborative Innovation Center of Intelligent Visual Computing, School of Computer Science, Fudan University, China\\
$^2$Tomorrow Advance Life, China\\
$^3$University of Science and Technology of China, China\\
\emails
tlzheng21@m.fudan.edu.cn,
\{zhinchen, ygj\}@fudan.edu.cn,
jfbai.bit@gmail.com,
htxie@ustc.edu.cn
}

\begin{document}

\maketitle

\begin{abstract}
Text irregularities pose significant challenges to scene text recognizers. Thin-Plate Spline (TPS)-based rectification is widely regarded as an effective means to deal with them. Currently, the calculation of TPS transformation parameters purely depends on the quality of regressed text borders. It ignores the text content and often leads to unsatisfactory rectified results for severely distorted text. In this work, we introduce TPS++, an attention-enhanced TPS transformation that incorporates the attention mechanism to text rectification for the first time. TPS++ formulates the parameter calculation as a joint process of foreground control point regression and content-based attention score estimation, which is computed by a dedicated designed gated-attention block. TPS++ builds a more flexible content-aware rectifier, generating a natural text correction that is easier to read by the subsequent recognizer. Moreover, TPS++ shares the feature backbone with the recognizer in part and implements the rectification at feature-level rather than image-level, incurring only a small overhead in terms of parameters and inference time. Experiments on public benchmarks show that TPS++ consistently improves the recognition and achieves state-of-the-art accuracy. Meanwhile, it generalizes well on different backbones and recognizers. Code is at \url{https://github.com/simplify23/TPS_PP}.

\end{abstract}

\section{Introduction}
Scene text recognition (STR) aims to understand and transcribe text images captured in the wild. With the prevalence of using cameras in applications such as autonomous driving, image retrieval, etc., it becomes one of the most active research themes nowadays \cite{Baekwhats_wrong_19ICCV,long2021str_era,Chen2021text,du2022@svtr,wang2022petr}. However, STR still faces the problem that the recognition accuracy is not satisfactory for difficult text, partly because the text is not presented and captured in canonical, resulting in distortions such as perspective, orientation, and curved text. While humans can accommodate these distortions, for example by focusing on discriminative character patterns, modern recognizers still struggle to recognize such instances. 

\begin{figure}[t]
\centering
\includegraphics[width=0.45\textwidth]{ 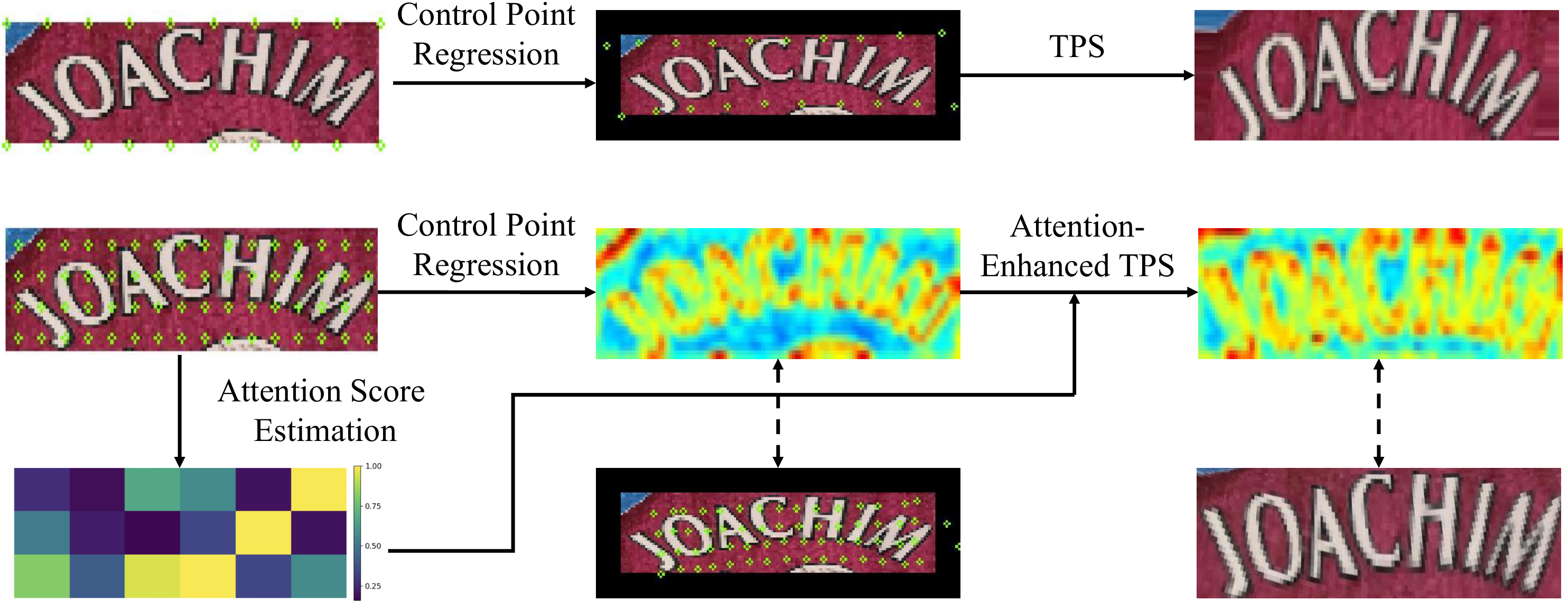} 
\caption{ASTER \protect\cite{shi2018aster} (the upper part) regresses control points from image borders to text borders. TPS transformation parameters are calculated based on the movements. TPS++ (the bottom part) initializes control points uniformly. The transformation parameters are computed based on both the predicted point movements and content-based attention scores. The rectification is conducted at the image feature-level. By projecting back to the image, we show the regressed control points (bottom center) and rectified image (bottom right). TPS++ gives a more natural text rectification.}

\label{fig0:motivation}
\end{figure}

Text rectification emerges as a promising way to relieve the problem. Typically, it serves as a pre-processing that aims to correct the distortions and generate a cropped image concentrating on text foreground and in near canonical form. Therefore STR can be simplified by applying to the rectified image. Early works like STN \cite{jaderberg2015stn,liu2016starnet} used affine transformation to correct the distortions from image translation, scale and rotation perspectives. The irregularities, however, are diverse and complicated, so only limited success was achieved. With research efforts accumulated, Shi et al. \cite{shi2016robust_auto_rect} formulated the rectification using Thin-Plate Spline (TPS) transformation \cite{bookstein1989tps}. As depicted in the upper part of Fig.\ref{fig0:motivation}, it first regresses a series of control points from image borders to text borders. Then, the movement of these points are used to calculate the TPS transformation parameters, which are further applied to cropping and rectifying the text. This branch of methods \cite{shi2016robust_auto_rect,shi2018aster,yang2019symmetry,qian2021adaptive} offers a flexible rectification that is able to deal with anisotropic local distortions. It is one of the most influential solutions to date. Meanwhile, it shows advantages such as the rectification being trained in conjunction with the recognition in a weakly supervised manner, where only text labels are required. Moreover, it is a plug-in that can be inserted into any recognizer.  

Despite great success, it is observed that this branch of rectifiers is less effective for difficult text. For example in Fig.\ref{fig0:motivation}, ASTER \cite{shi2018aster} does not get a satisfactory rectification. The main reason lies in that it determines the TPS transformation parameters according to the regressed text borders, which are obtained via weakly supervised learning and are sometimes not well localized. More importantly, the borders carry very little text information. The transformation has not been taught how to accommodate characteristics of the text. Therefore, the rectified image, although basically maintaining the text's geometric structure, also gives rise to unnatural character deformation and out of bounds (e.g., the upper part of character "C" in Fig.\ref{fig0:motivation}). On the other hand, there are a few studies that employ pipelines other than text border regression. For example, MORAN \cite{cluo2019moran}, ESIR \cite{zhan2019esir}, STAN \cite{lin2021stan}, etc. However, they either run slowly or do not perform better than the border-regression-based methods.

One essential cause of unnatural rectification is that the calculation of TPS parameters is content-free. The transformation is performed regardless of the text content. To solve it, more flexibility should be given to the computation of TPS parameters, allowing the correction to be aware of the text. It is therefore capable of suppressing undesired deformations by using text content to constrain the control point movement. Meanwhile, it is observed that the vast majority of methods treat rectification as a pre-processing. Therefore, image features are extracted independently for both rectification and recognition, wasting computational resources considerably. A lightweight and fast rectifier is supposed to obtain if it can share some features in common with the recognition. 

With these observations, we propose TPS++, an attention-enhanced TPS to address the above issues. TPS++ aims to leverage the attention mechanism to establish correlations between control points and text content. By integrating them into the TPS formula, it assigns extra flexibility to the transformation, generating a more natural rectification that benefits the recognition. To enable this, three major upgrades are made. First, we re-endow the role of control points. Unlike previous studies that initiate them along the image borders, TPS++ adopts a grid-like initialization that uniformly distributes them spatially. Besides using their movements to estimate the TPS parameters, a majority of control points are located in text foreground rather than text borders. Therefore they catch the text content which is required in attention modeling. Second, we develop a gated-attention mechanism dedicated to modeling attention scores between control points and text content. With the score, additional flexibility is injected into the transformation. We devise a new formula that takes both the movement of control points and attention scores into account, building a more flexible content-aware TPS. Third, TPS++ gives a new paradigm where the rectifier is coupled with the recognizer. The rectifier shares the feature backbone with the recognizer in part. We perform the rectification at the image feature-level. The control point prediction and attention score estimation are carried out on feature maps that are also more flexible. It leads to a lightweight and efficient implementation. Meanwhile, it also has the merits of the two tasks easier to be jointly optimized. Extensive experiments are conducted on public benchmarks to verify the effectiveness of TPS++. It shows that adding TPS++ consistently improves the recognition accuracy. It correctly recognizes difficult text instances and achieves state-of-the-art accuracy, while the introduced overhead on parameters and inference time is just 0.5M and a few milliseconds. Moreover, TPS++ generalizes well on different backbones and recognizers.

TPS++ has four appealing properties. First, a flexible rectifier. It adds the attention score to TPS transformation, giving a more flexible text correction. To our knowledge, this is the first work exploiting the attention mechanism in text rectification. Second, an accurate rectifier. It gives large performance gains and achieves state-of-the-art accuracy when applied to popular recognizers. Third, a lightweight rectifier. It performs the rectification at feature-level instead of image-level, thus introducing very little overhead in terms of parameters and inference speed by allowing feature sharing. Fourth, a universal rectifier. It can be embedded into different backbones and recognizers without or with only trivial modifications, and consistently yielding accuracy improvements.

\section{Related Work}
\subsection{Scene Text Recognition}
STR is an intensively studied task in the deep learning era. Sequence-based models became popular due to their ability in integrating recognition clues from different aspects, e.g., visual, semantic, linguistic. Initially, CRNN \cite{ShiBY17crnn} encoded the input image as a visual feature sequence, which was then modeled by BiLSTM for context reinforcement and CTC loss for text transcription. The paradigm was extended to GTC \cite{hu2020gtc} by incorporating graph neural network and attention mechanism, and SVTR \cite{du2022@svtr} by leveraging visual transformers (ViTs). 
To use semantic clue, attention-based encoder-decoder models \cite{lee2016attention_origin,cheng2017focusing_attention,fang2018attention,sheng2019nrtr,bhunia2021jvsr} injected features extracted from text labels to the decoder, exploring the way of utilizing clues from both visual and semantic aspects. Robustscanner \cite{yue2020robustscanner} and CDistNet \cite{zheng2021cdistnet} devised dedicated position modeling modules, which were helpful in decoding characters by using the correct image features. Nevertheless, these methods used an iterative decoding scheme that identified the characters one-by-one, resulting in a slow speed. Consequently, parallel decoding schemes such as SRN \cite{SRNyu2020towards}, VisionLAN \cite{wang2021FTO}, ABINet \cite{ABInet21CVPR} were introduced. Character placeholders were appropriately initialized such that the characters could be estimated in parallel with accelerated speed. Meanwhile, the linguistic clue was also modeled to maintain accuracy. Recently, self-supervised-based pre-training was taken into account in STR and led to improved accuracy \cite{yang2022reading,yu2023structextv2}.

While these models gained steady performance improvements, they did not specifically take into account text irregularities, thus less effective for irregular text. To overcome this problem, segmentation-based and rectification-based methods were developed. The former regarded the recognition as a character-level segmentation task and exhibited impressive performance in dealing with these distortions \cite{liu2018charnet,liao2019two_dim_per,xing2019convolutional,li2017fcsem}. Nevertheless, they required character-level annotations which were not always readily available. Moreover, the segmentation was also sensitive to noise. On the other hand, rectification-based methods aim to rectify the text to generate a near canonical and easier readable counterpart. They were compatible with most off-the-shelf recognizers and thus received considerable research attention.

\subsection{Text Rectification}
Based on the transformation formulation, we roughly classify rectification-based methods into affine transformation-based, TPS-based methods and others.
The first aims to eliminate the distortion based on simple transformations such as translation, scale and rotation. The first work was STN \cite{jaderberg2015stn} proposed for general object rectification. It was studied in STR by applying the transformation on characters \cite{liu2016starnet} and image patches \cite{lin2021stan}.
TPS-based approaches noted that the text captured in the wild exhibits anisotropic distortions, which were difficult to describe by affine deformations. Shi et al. proposed RARE \cite{shi2016robust_auto_rect} and ASTER \cite{shi2018aster} that enabled more flexible rectification by using TPS \cite{bookstein1989tps} transformation instead. However, their weakly supervised nature makes the calculation of TPS parameters less accurate for highly distorted text.
To better handle such instances, an iterative rectified scheme was developed in \cite{zhan2019esir}. It reduced the dependence on point regression quality but increased the time consumption. In \cite{yang2019symmetry}, a more accurate text border regression was obtained by predicting a series of text geometric attributes and utilizing them to assist the rectification. However, costly character-level supervision was required. Besides the two major branches, MORAN \cite{cluo2019moran} regarded text image as a grid and directly regressed offsets at each cross point. However, unnatural distortion was observed due to these offsets are not constrained by any geometric transformation. Zhang et al. \cite{zhang2021spin} paid attention to color-related rectification and devised a structure-preserving rectifier.

Compared with these methods, TPS++ makes use of the attention mechanism to improve the quality of rectification. By adding attentional parameters, it greatly alleviates unnatural distortions observed in current rectifiers such as TPS. Moreover, it performs the rectification at feature-level rather than image-level, thus with only a limited recognition overhead in terms of parameters and inference speed.

\begin{figure*}[t]
\centering
\includegraphics[width=0.85\textwidth,height=0.4\textheight]{ 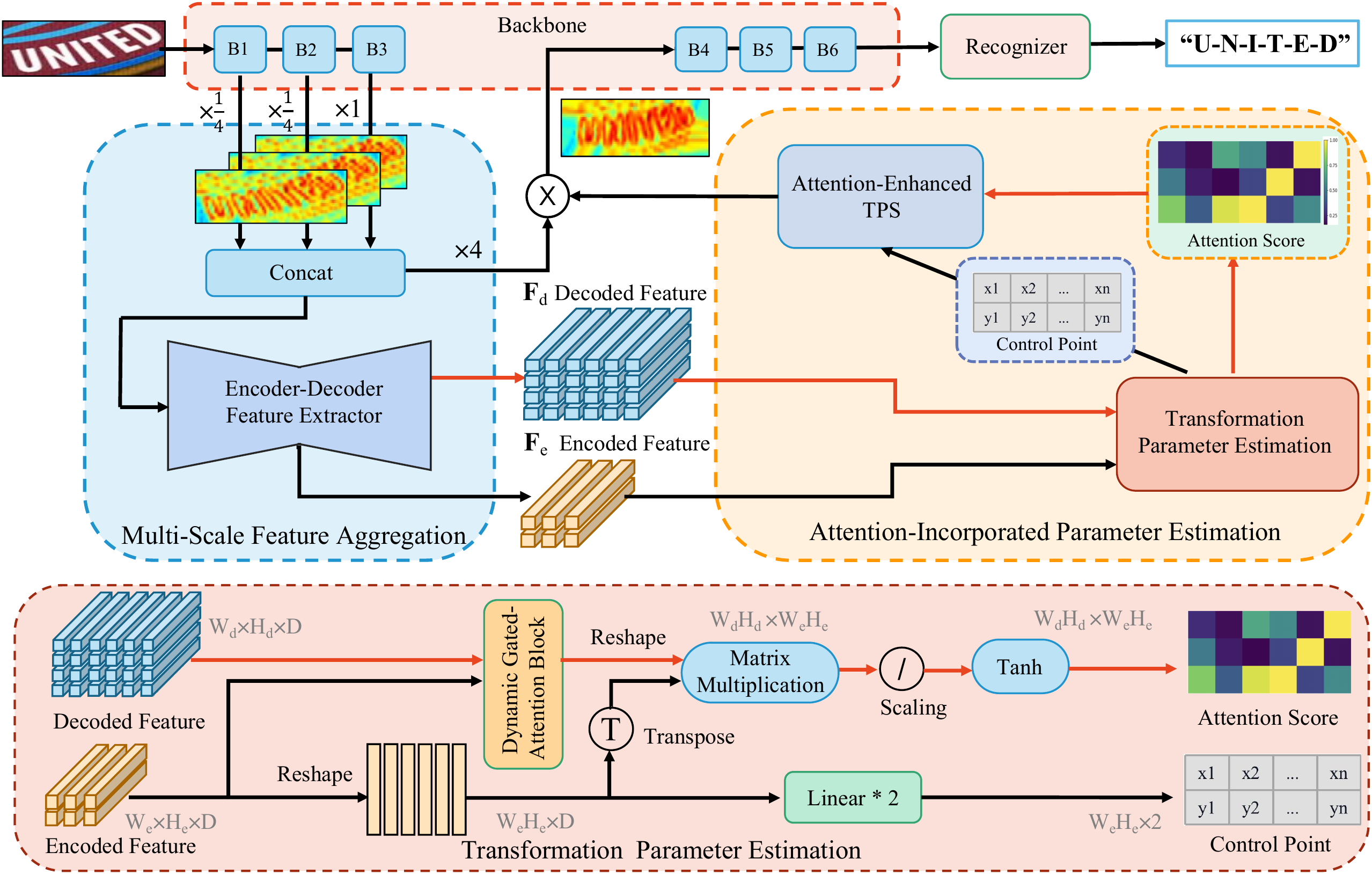} 
\caption{An illustrative framework of TPS++. It consists of two parts: Multi-Scale Feature Aggregation (MSFA) and Attention-Incorporated Parameter Estimation (AIPE), respective for visual feature aggregation and attention-enhanced TPS parameter estimation.}
\label{fig1:TPS++}
\end{figure*}

\section{Methodology}
\subsection{Overview}
An illustrative framework of TPS++ is depicted in Fig.\ref{fig1:TPS++}, which consists of two parts: multi-scale feature aggregation (MSFA) and attention-incorporated parameter estimation (AIPE). We taking ResNet-45 as the backbone for illustration. Given a text instance, feature maps of the first three blocks in the backbone are injected into MSFA. These features are aggregated and two outputs are produced, i.e., the encoded feature $\boldsymbol{F}_{e}$ and decoded feature $\boldsymbol{F}_{d}$. Then, AIPE uses the two features to predict the movement of control points and attention matrix $\boldsymbol{A}$, which records the attention score between control points and text content. Based on these parameters, attention-enhanced TPS rectification is performed at the image feature-level. The rectified features are fed back to the backbone. The subsequent feature extraction and recognition are carried out the same as conventional. 

TPS++ is featured by two distinct properties compared with existing rectifiers. First, it shares visual feature extractor in part between recognition and rectification, generating a tight coupling scheme that well controls the parameter and inference speed overhead, while also largely preserving its plug-in nature. Second, it introduces the attention mechanism to TPS, enabling a more flexible content-aware correction. Both improve the rectification quality and ease the recognition. Meanwhile, TPS++ also inherits merits such as end-to-end trainable with STR, requiring no extra annotations beyond text labels. Details will be elaborated later.

\begin{table}[]
\centering
\resizebox{0.7\linewidth}{!}{
\begin{tabular}{|c|c|c|}
\hline
\multirow{1}{*}{Layers} & \multirow{1}{*}{Configuration} & \multirow{1}{*}{Output} \\
\hline
Layer1                  & Conv(192,1*1,1)              & 16*64                   \\ \hline
Layer2                  & Conv(64,3*3,2)               & 8*32                    \\ \hline
Layer3                  & Conv(64,3*3,2)               & 4*16                    \\ \hline
Layer4                  & CBAM~\cite{woo2018cbam}                         & 4*16                    \\ \hline
Layer5                  & Up-Conv(64,3*3,2)               & 8*32                   \\ \hline
Layer6                  & Up-Conv(64,3*3,2)               & 16*64                    \\ \hline
Layer7                  & 
Conv(64,3*3,1)               & 16*64                   \\ \hline
\end{tabular}}
\caption{Structure of the encoder-decoder feature extractor.}
\label{feature_extractor}
\end{table}

\subsection{Multi-Scale Feature Aggregation}
Most previous text rectifiers \cite{shi2016robust_auto_rect,shi2018aster} are designed as a pre-processing before STR. The scheme is computationally intensive as feature extraction is executed twice. To overcome the drawback, TPS++ shares the feature backbone in part with the recognizer. We design a thin module called MSFA that accepts feature maps generated from the first three backbone blocks. The features are scaled to the same size and concatenated from the channel dimension. Specifically, features from the 1st and 2th blocks are reduced by $4\times$ spatially, while feature channels of the 1st, 2th and 3rd blocks are all aligned to 64. Then, a lightweight encoder-decoder-based feature extractor, whose structure is provided in Tab.\ref{feature_extractor}, is applied. It consists of a contracting path and a symmetric expansive path, both containing three convolution layers. In addition, CBAM \cite{woo2018cbam}, the channel-spatial joint attention, is applied to highlight important features. With these operations, multi-scale visual features are aggregated and optimized towards the rectification purpose. Features obtained from CBAM (i.e., $\boldsymbol{F}_{e} \in R^{W_{e}\times H_{e}\times D}$) and the outputted layer (i.e., $\boldsymbol{F}_{d} \in R^{W_{d}\times H_{d}\times D}$) are termed as the encoded and decoded features, respectively. Note that the two features have the same number of channels and $\boldsymbol{F}_{d}$ has the same spatial resolution as the scaled input feature.

Separating the feature extraction into two parts also is a key factor that makes TPS++ effective. By doing this, the first part emphasizes extracting generic visual features. While the second is MSFA which targets rectification optimization. It aggregates features from shallow blocks that contain more location-related clues, which is useful for control points regression and attention modeling. Note that the second part is slim and introduces only a few computational overhead.

\subsection{Attention-Incorporated Parameter Estimation}

\begin{figure}[t]
\centering
\includegraphics[width=0.45\textwidth,]{ 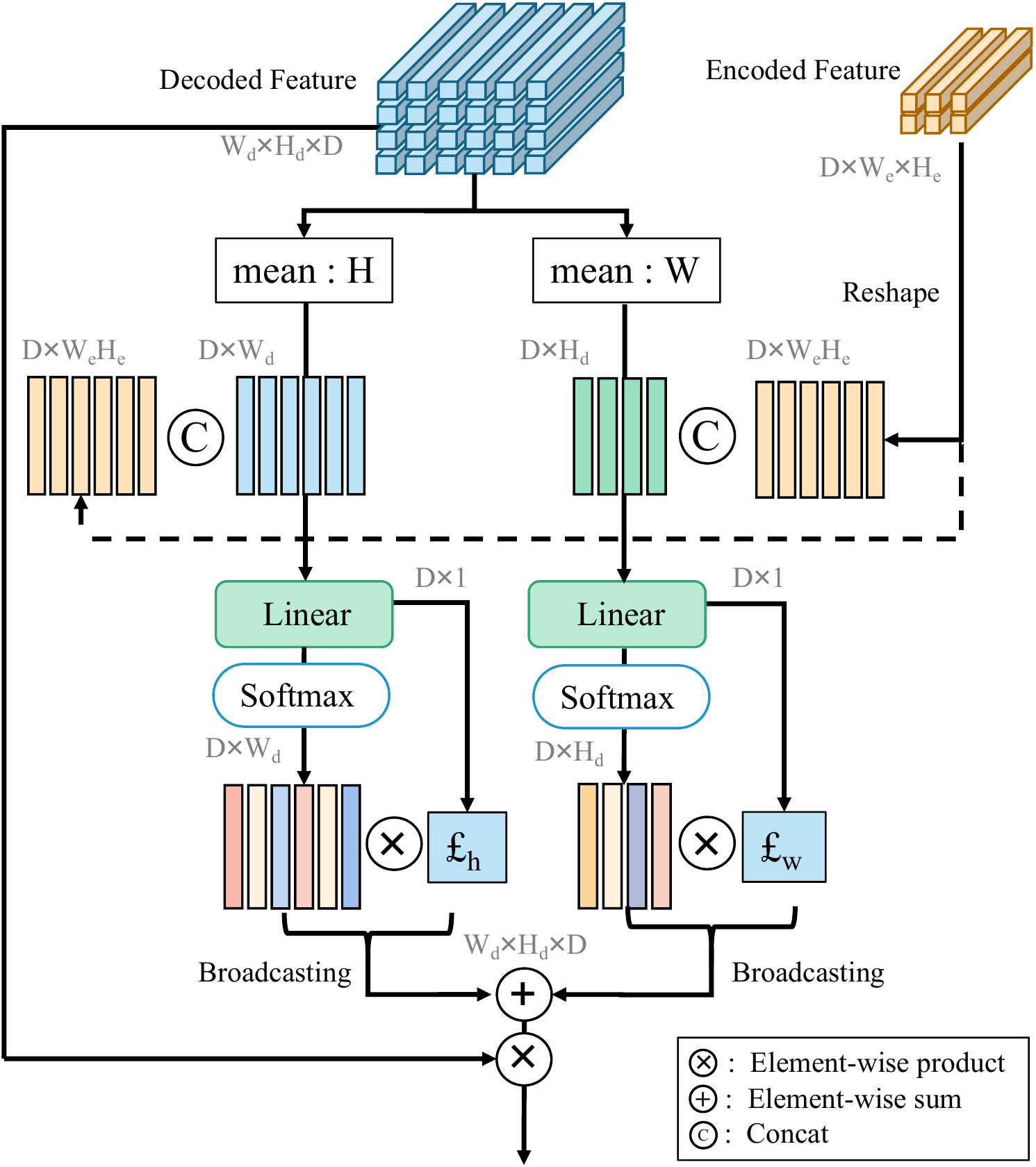} 
\caption{The detail of dynamic gated-attention block (DGAB).}
\label{fig1:motivation}
\end{figure}

AIPE is appended behind MSFA for control point regression and content-based attention score estimation. As depicted in the middle right of Fig.\ref{fig1:TPS++}, besides regressing control points, it predicts a content-aware attention score matrix $\boldsymbol{A}$ in parallel by using the gated-attention mechanism. Specifically, in control point regression, unlike TPS initializes the points along the image borders, we set control points ${\boldsymbol{C}} \in R^{W_{e}H_{e}\times 2}$ uniformly distributed on the feature map spatially, i.e., a gird-like distribution. Therefore, a large portion of points is located in text foreground rather than less informative borders. Note that the number of control points equals the spatial resolution of $\boldsymbol{F}_{e}$. Meanwhile, $\boldsymbol{F}_{e}$ is reshaped as a feature sequence $\boldsymbol{\hat{F}}_{e} \in R^{W_{e}H_{e}\times D}$. It undergoes two linear layers to predict the offsets of each control point on \emph{x} and \emph{y} dimensions. Combining with the initialized coordinates, we get a set of regressed control points ${\boldsymbol{C}^\prime} \in R^{W_{e}H_{e}\times 2}$.

In attention score estimation, we assess the correlation between $\boldsymbol{F}_{e}$ and $\boldsymbol{F}_{d}$ to get an attention score matrix $\boldsymbol{A}$ that is aware of text content. Specifically, the two features both feed into a dedicated designed module called dynamic gated-attention block (DGAB), where attention scores between control points and text are adaptively computed. Then, the obtained feature is reshaped and further combined with $\boldsymbol{\hat{F}}_{e}$ via matrix multiplication, followed by a scaling of $1/\sqrt{D}$ and Tanh activation that restricts the attention scores to (-1, 1).

Fig.\ref{fig1:motivation} gives the detail of DGAB. First, the decoded feature $\boldsymbol{F}_{d}$ is shrunk to two feature sequences of $D \times W_{d}$ and $D \times H_{d}$ by averaging along the $H$ and $W$ dimensions, respectively. Then, the two sequences are separately concatenated with the reshaped encoded feature $\boldsymbol{\hat{F}}_{e}$, followed by the linear layer to align the dimension back to $D \times W_{d}$ and $D \times H_{d}$ and dynamically calculate their weights along $W_{d}$ and $H_{d}$ dimensions, namely £$_{h}$ and £$_{w}$. They represent the importance of each column and row, respectively. Meanwhile, the aligned features experience a Softmax operation and are multiplied by the derived weights. In the following, the obtained features are broadcasted back to dimension $W_{d}\times H_{d}\times D$ and are merged by element-wise sum. An element-wise product with the raw $\boldsymbol{F}_{d}$ is followed up to generate the output. 

With the estimated control points and attention scores, we develop a more flexible attention-enhanced TPS transformation. For each location $p_{i}^{\prime}$ in the corrected feature map, its original location $p_{i}$ is determined by a geometric transformation as given by Equ.\ref{eq31}. Then, $p_{i}$ is computed by $\boldsymbol{T}$ and $\boldsymbol{F}(\cdot)$ contains the attention score as follows.

\begin{equation}
p_{i}=\boldsymbol{T} \cdot \boldsymbol{F}\left(p_{i}^{\prime}\right)
\label{eq31}
\end{equation}
\begin{gather}
\boldsymbol{T}=\left(\left[\begin{array}{lll}
1_{K} & \boldsymbol{C}^{\prime T} & \boldsymbol{S} \\ \\
0 & 0 & 1_{K}^{T} \\ \\
0 & 0 & \boldsymbol{C}^{\prime}
\end{array}\right]^{-1}\left[\begin{array}{l}
\boldsymbol{C}^{T} \\ \\
0 \\ \\
0
\end{array}\right]\right)^{T}
\label{eq4}
\end{gather}
\noindent where $K=W_{e}H_{e}$ is the number of control point. $\boldsymbol{S} \in R^{K\times K}$ is a square matrix with element $s_{ij} = E_{u}(||c_{i}-c_{j}||)$ defined as radial basis kernel applied to the Euclidean distance between $c_{i} \in \boldsymbol{C}$ and $c_{j} \in \boldsymbol{C}$.

\begin{gather}
\boldsymbol{F}\left(p^{\prime}_{i}\right)=\left[\begin{array}{c}
1 \\
p^{\prime}_{i} \\
E_{u}\left(\left\|p^{\prime}_{i}-c^{\prime}_{1}\right\|\right) *\left(\lambda \cdot \boldsymbol{A}_{i,1}+\beta\right) \\
\vdots \\
E_{u}\left(\left\|p^{\prime}_{i}-c^{\prime}_{k}\right\|\right) *\left(\lambda \cdot \boldsymbol{A}_{i,k}+\beta\right)
\end{array}\right]
\label{eq3}
\end{gather}

\noindent where $\boldsymbol{A}_{i,k}$ is the attention score between the $i$-th location and the $k$-th control point. $\lambda$ and $\beta$ are hyperparameters empirically set to 0.5 and 1, respectively. When $\lambda$=0, the equation goes back to conventional TPS exactly.

As seen in Equ.\ref{eq3}, TPS++ gives rise to additional flexibility to TPS by incorporating the attention score. It allows a content-aware adaptive weighting on control points when performing the transformation. Since the rectification and recognition are jointly optimized, this flexibility is able to guide parameter updated towards a better STR, and meanwhile, generating a more natural rectification. Note that several studies \cite{lee2016attention_origin,cheng2017focusing_attention,shi2018aster,sheng2019nrtr,lin2021stan} also incorporated the attention mechanism in STR, but their attention is considered in the text recognition stage only. To the best of our knowledge, TPS++ is the first that introduces the attention mechanism to text rectification.

\subsection{TPS++ on Different Recognizers}
One appealing property of TPS is the plug-in nature. It can be seamlessly inserted into any text recognizer. TPS++ largely preserves this property. In case ResNet-45 is employed as the backbone, TPS++ can be directly appended no matter which recognizer is employed, otherwise a modification might be required. To be compatible with MSFA, we should give three feature maps whose sizes can be normalized to $16\times64\times64$. This is not a strict restriction for popular CNNs even ViTs. Therefore to apply TPS++, only trivial accommodation is required to align the features in spatial and channel. We will verify this capability of TPS++ in experiments.


\section{Experiments}

\subsection{Datasets and Implementation Details}
Following the standard protocol in STR \cite{Baekwhats_wrong_19ICCV}, models are trained on two synthetic datasets and evaluated on six public benchmarks, which are as follows:

\noindent\textbf{MJSynth (MJ)} \cite{MJor90K} and \textbf{SynthText (ST)} \cite{ST} are the two synthetic datasets with 8.91M and 6.95M text instances, respectively.


\noindent\textbf{ICDAR2013 (IC13)} \cite{ICDAR2013}, \textbf{Street View Text (SVT)} \cite{SVT}, \textbf{IIIT5k-Words (IIITK)} \cite{IIIT5K}, \textbf{ICDAR2015 (IC15)} \cite{ICDAR2015}, \textbf{SVT-Perspective (SVTP)} \cite{SVT-P} and \textbf{CUTE80 (CT80)} \cite{CUTE80} are six benchmarks widely used in assessing STR models. The first three mainly contain regular text while the rest three are irregular.

All models were trained with Adam optimizer for 12 epochs on the two synthetic datasets, only word-level annotations are utilized. The initial learning rate was set to $1e^{-3}$, which was reduced to $1e^{-4}$ and $1e^{-5}$ at the 8th and 10th epoch, respectively. All input images were resized to $32\times128$. The batch size was set to 200. Warm-up strategy was used in the first epoch, and the initial warm-up ratio was set to 0.001. For different recognizers, for fairness we follow the parameter settings as their papers and performed data augmentation uniformly. TPS++ was trained by two steps. The recognizer was trained w/o TPS++ at first. Then, TPS++ was appended and jointly trained again. All models were trained by using a server with 6 NVIDIA 3080 GPUs.

\subsection{Ablation Study}
We employ ABINet-V \cite{ABInet21CVPR} as the base network. It uses ResNet-45 and two Transformer units for STR. The language model is not taken into account. 

\noindent\textbf{Backbone feature utilization.}
Since TPS++ is coupled with recognition, how to share the backbone becomes an issue. We perform experiments to ablate this. The first feeds the raw image directly to the encoder-decoder-based feature extractor in MSFA (i.e., with insert position 0. Below defined similarity). It treats rectification and recognition separately as conventional methods. While the second and third trails acquire feature maps behind the 3rd and 6th blocks, respectively. As seen in Tab.\ref{tab:rec_pos}, the performance decreases for higher insert positions. It is explained as the feature spatial resolution is reduced along with the increase of insert position, thus less information is carried. For example, the first scheme operates on the full-resolution image while the second on a $16\times$ smaller one. To tackle this dilemma, we propose to aggregate multi-scale features and scale them to the same size as features from the 3rd block. It not only gets the best performance but also the computational cost is well controlled.

\begin{table}[t]
\resizebox{1.0\linewidth}{!}{
\begin{tabular}{ccccccccc}
\hline
\multirow{2}{*}{Insert Pos.}  & \multirow{2}{*}{Multi-scale}  & \multirow{2}{*}{IIIT5k} & \multirow{2}{*}{SVT}  & \multirow{2}{*}{IC13} & \multirow{2}{*}{IC15} & \multirow{2}{*}{SVTP}  & \multirow{2}{*}{CUTE}   \\ 
\\
\hline
0 & & 94.1 & 90.0 & 91.8 & 77.2 & 80.8 & 83.1 \\
3 & & 93.4 & 89.2 & 91.9 & 77.0 & 81.7 & 82.6\\
6 & & 91.5 & 87.3 &90.1 & 73.1 & 78.6 & 79.9\\
3 &  \checkmark &94.1 & 91.2 & 92.2 & 78.7 & 82.6 & 84.3\\
\hline
\end{tabular}
}
\caption{Ablation study on backbone feature utilization.}
\label{tab:rec_pos}
\end{table}

\noindent\textbf{Number of control points.} 
We assess how the accuracy and inference time varies with the number of control points. As shown in Tab.\ref{tab:point_number}, increasing the number of control points along either width or height dimension both improves the recognition accuracy. It achieves the highest accuracy when $4\times16$ points are sampled. This is not surprising as sampling denser also enables a fine-grained perception of text content, therefore better alleviating the problem of unnatural character deformations. On the other hand, it is also seen that different sampling schemes have very close inference time consumption, indicating that the control point-related computation is efficient. Therefore the $4\times16$ scheme is chosen. 

\begin{table}[t]
\resizebox{1.0\linewidth}{!}{
\begin{tabular}{ccccccc|c}
\hline
\multirow{2}{*}{Point Num.}  & \multirow{2}{*}{IIIT5k} & \multirow{2}{*}{SVT}  & \multirow{2}{*}{IC13} & \multirow{2}{*}{IC15} & \multirow{2}{*}{SVTP} & \multirow{2}{*}{CUTE} &  Time \\ 
&&&&&&&(ms)
\\
\hline
2$\times$ 4 & 93.4 & 89.2 & 91.9 & 77.0 & 81.7 & 81.6 &  17.3\\
2$\times$ 8 & 93.8 & 90.1 & 92.1 & 78.3 & 82.3 & 82.6 & 17.5\\
4$\times$ 8 & 94.2 & 90.9 & 91.9 & 78.5 & 82.3 & 83.3  & 17.5\\
8$\times$ 16 &94.0 & 91.2 & 92.0 & 78.1 & 82.6 & 83.3  & 17.6\\
\hline
4$\times$ 16 &94.1 & 91.2 & 92.2 & 78.7 & 82.6 & 84.3  & 17.5\\

\hline
\end{tabular}}

\caption{Ablation study on number of control points.}
\label{tab:point_number}
\end{table}

\begin{table}[t]
\resizebox{1.0\linewidth}{!}{
\begin{tabular}{c|c|cccccc|c}
\hline
\multirow{2}{*}{Rect based} & \multirow{2}{*}{Atten}  & \multirow{2}{*}{IIIT5k} & \multirow{2}{*}{SVT}  & \multirow{2}{*}{IC13} & \multirow{2}{*}{IC15} & \multirow{2}{*}{SVTP} & \multirow{2}{*}{CUTE}  &\multirow{1}{*}{+Params} \\
&&&&&&& & ($\times 10^{6}$)
\\ 
\hline
Baseline  & -- &91.5 &87.8 &90.1 &73.5 &78.6 & 80.6 & -- \\ 
MORAN\shortcite{cluo2019moran} & -- &93.0 & 87.5 & 90.9 & 74.3 & 79.1 &82.6  & 0.2\\ 
ASTER\shortcite{shi2018aster} & -- & 93.5 & 88.9 & 92.3 & 76.7 & 80.5 & 84.7 & 1.7 \\
SPIN\shortcite{zhang2021spin} & -- & 93.7 &89.3 &91.3 &78.2 &78.8 &83.0  & 2.3 \\ 
\hline
Grid  & --  & 93.7 & 89.6 & 91.1 & 76.2 & 80.6 & 82.6  & 0.3\\
Gird+Atten & -- & 93.7 & 89.8 & 92.5 & 77.6 & 81.9 & 83.3 & 0.5\\
TPS++ & w & 94.1 & 91.2 & 92.2 & 78.1 & 82.6 & 84.3 & 0.4 \\
TPS++ & w+h & 94.5 & 91.5 & 92.8 & 78.2 & 82.8 & 85.8 & 0.5 \\
\hline
\end{tabular}
}
\caption{Comparison on rectifiers and TPS++ components.}
\label{tab:attention}
\end{table}

\begin{figure}[t]
\centering
\includegraphics[width=0.45\textwidth]{ 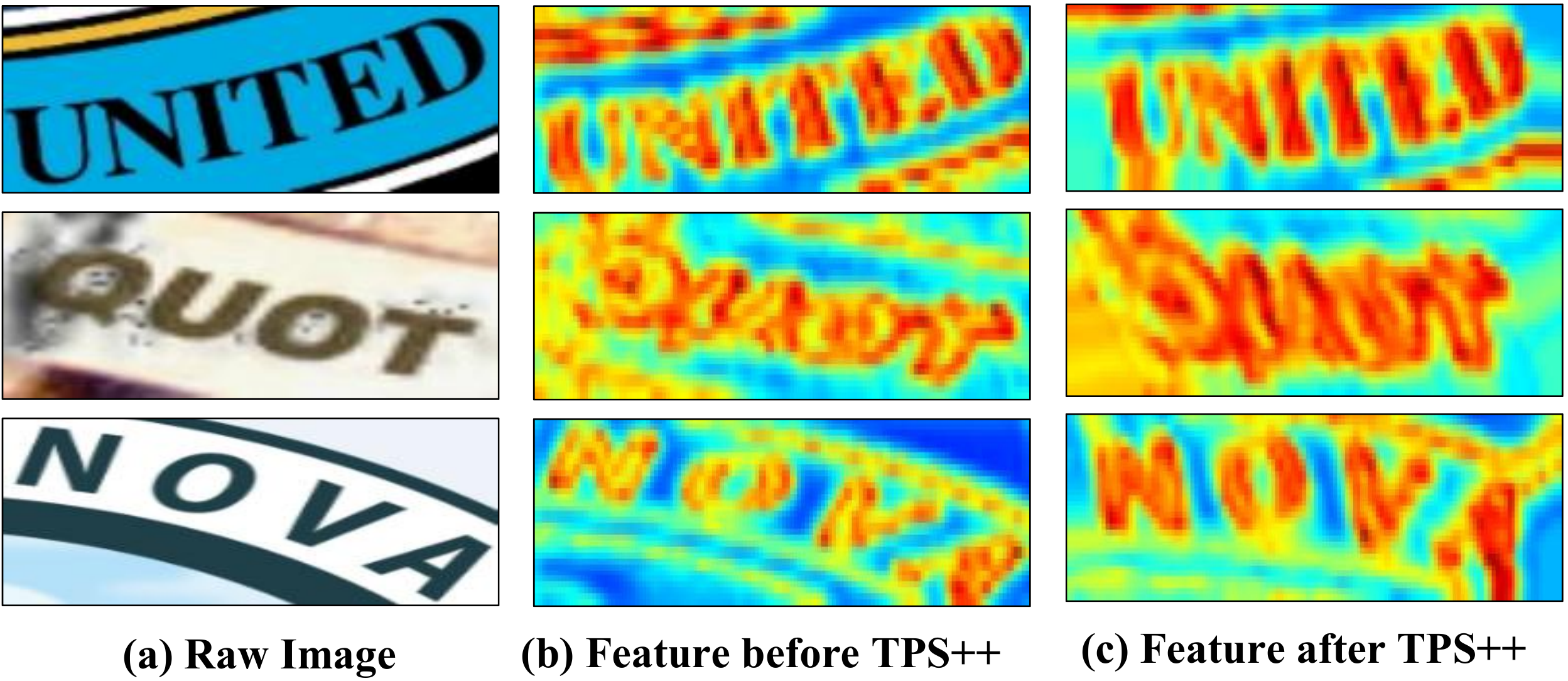} 
\caption{Visual feature visualization before and after TPS++.}
\label{fig1:vis_feature}
\end{figure}
\begin{figure}[t]
\centering
\includegraphics[width=0.5\textwidth]{ 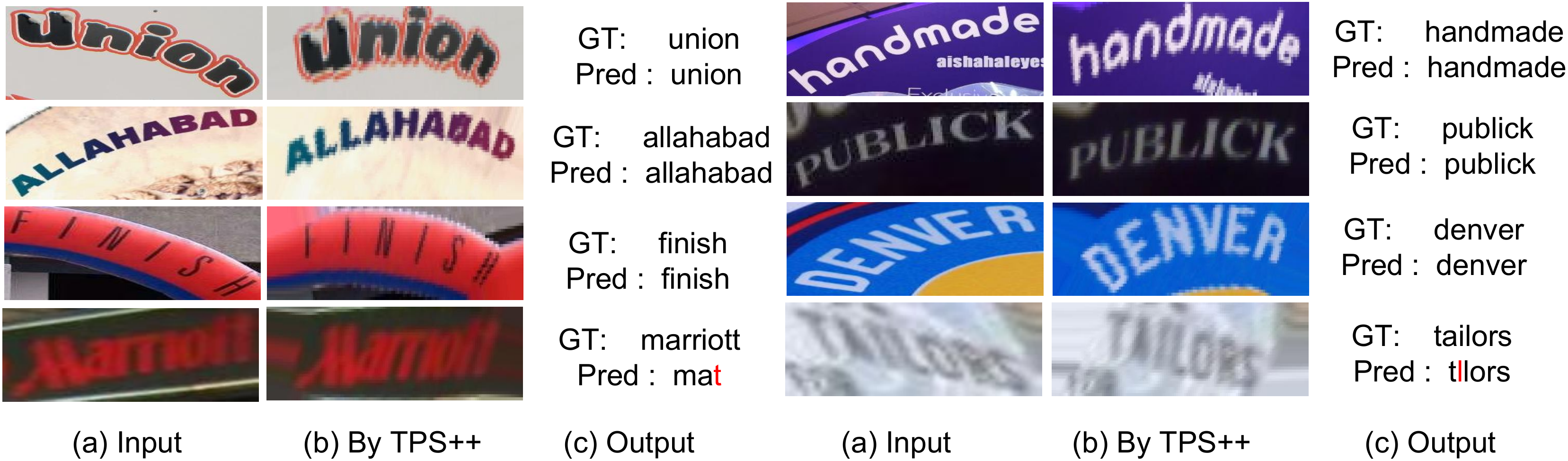} 
\caption{Good and bad cases produced by TPS++.}
\label{fig:vis_point}
\end{figure}

\begin{table*}[h]
\centering
\begin{tabular}{c|c|c|c|c|c|c|c|c|c}
	\hline
	\multirow{2}{*}{Methods} & \multirow{2}{*}{Training Data}   & \multicolumn{3}{c}{Regular} & \multicolumn{3}{|c|}{Irregular} & \multirow{1}{*}{Params} &\multirow{1}{*}{Time}  \\
	\cline{3-8}
	&  &IIIT5k & SVT  & IC13 &  IC15 & SVTP & CUTE & ($\times 10^{6}$) & (ms)  \\
	\hline
CRNN\cite{ShiBY17crnn}  & 90K & 78.2  & 80.8 & 86.7 & -- & --& -- & 8.3 & 6.8 \\
SAR\cite{li2019sar} & 90K+ST & 91.5 & 84.5 & 91.0 & 69.2 & 76.4 & 83.3 & 57.5 & 120\\
NRTR\cite{sheng2019nrtr}  & 90K+ST & 90.1 & 91.5 & 95.8 & 79.4 & 86.6 & 80.9 & 31.7 & 212\\
RobustScanner\cite{yue2020robustscanner} & 90K+ST  & 95.3 & 88.1 & 94.8 & 77.1 & 79.5 & 90.3 & -- &--\\
SRN\cite{SRNyu2020towards}  & 90K+ST & 94.8 & 91.5 & 95.5 & 82.7 & 85.1 & 87.8 & 49.3 & 26.9\\
Pren2D\cite{yan2021primitive} & 90K+ST+Real & 95.6 & 94.0 & 96.4& 83.0 & 87.6& 91.7 & -- & 67.4\\
VisionLAN\cite{wang2021FTO} & 90K+ST  & 95.8& 91.7& 95.7 & 83.7 & 86.0 & 88.5 & 32.8 & 28.0\\
ABINet-LV\cite{ABInet21CVPR}  & 90K+ST & 96.2 &93.5& 97.4& 86.0 &89.3&89.2 & 36.7 & 37.2\\
GTR\cite{he2021S-GTR} & 90K+ST & 95.8 & 94.1 & 96.8 & 84.6 & 87.9 &92.3 & 42.1 & 18.8 \\
PARSeq\cite{bautista2022parseq}& 90K+ST &\textbf{97.0} &93.6 &  97.0 & \textbf{86.5} & 88.9 & 92.2& 23.8 & 11.8\\
SGBANet\cite{zhong2022sgbanet} & 90K+ST & 95.4 & 89.1 & 95.1 & 78.4 & 83.1 & 88.2  & -- &--\\

\hline
ASTER\cite{shi2018aster} & 90K+ST  & 93.4 & 89.5 & 91.8 & 76.1 & 78.5 & 79.5 & 22 & 73.1\\
ESIR\cite{zhan2019esir} & 90K+ST  & 93.3  & 90.2 & 91.3 & 76.9 & 79.6 & 83.3 & -- & -- \\
MORAN\cite{cluo2019moran} & 90K+ST & 91.2 & 88.3 & 92.4 & 68.8 & 76.1 & 77.4 & 28.5 & 24.4\\
STAN\cite{lin2021stan} & 90K+ST & 94.1 & 90.6 & 92.8 & 76.7 & 82.2 & 83.3 & -- & --\\
SPIN\cite{zhang2021spin} & 90K+ST & 95.2 & 90.9 & 94.8 & 79.5 & 83.2 & 87.5 & -- & --\\
\hline
CRNN+TPS++ & 90K+ST &  93.3 & 89.8 & 92.8 & 80.3 & 80.6 & 85.4  &  16.2 & 16.6\\ 
NRTR+TPS++ & 90K+ST & 96.3 & \textbf{94.6} & 96.6 & 85.7 & 89.0 & \textbf{92.4} & 35.5 & 218 \\ 
ABINet-LV+TPS++ & 90K+ST & 96.3 & 94.3 & \textbf{97.8} & \textbf{86.5} & \textbf{89.6} & 89.6 & 37.2 & 41.5 \\ 
	\hline
\end{tabular}

\caption{Performance comparison on six standard benchmarks. The accuracy of existing methods comes from their papers. While Params and Time are our reproduction using the same hardware.}
\label{tab:sota}
\end{table*}

\noindent\textbf{Attention formulation.}
We evaluate different attention formulations in TPS++ and compare them with existing rectification models. To maintain fairness, the backbone and recognition network are kept consistent. The results are presented in Tab.\ref{tab:attention}, where Baseline does not take into account the rectification. Gird denotes that only $\boldsymbol{F}_{e}$ is utilized to regress control points. Grid+Atten means a simple attention mechanism is considered (w/o utilizing DGAB). TPS++ (w) means that DGAB is only considered in the width dimension. MORAN \cite{cluo2019moran}, SPIN \cite{zhang2021spin} and ASTER \cite{shi2018aster} are popular rectifiers and we reproduce them using ABINet-V, a more powerful recognizer.

As seen, MORAN employs an image-level grid rectification. It performs worse than Gird although improvements are observed compared with Baseline, indicating the effectiveness of performing the rectification at feature-level. ASTER uses TPS transformation to perform the rectification. The better result shows that anisotropic transformation is vital for text irregularities correction. Grid+Atten gets accuracy advantages on 5 out of the 6 benchmarks compared with ASTER, indicating the efficacy of incorporating the attention mechanism which brings more flexibility. While TPS++ further employs a dedicated gated-attention along both the width and height dimensions to reinforce the attention modeling. The accuracy increases step-by-step. It achieves the best accuracy in all 6 benchmarks while the parameter increment is 0.5M, much less than the 1.7M in ASTER \cite{shi2018aster} and 2.3M in SPIN \cite{zhang2021spin}. In Fig.\ref{fig1:vis_feature}, three exemplars with their feature attention heatmaps before and after TPS++ are visualized, which is obtained by averaging feature activation along the channel dimension. It shows the feature has been rectified considerably after TPS++. Meanwhile, text foreground is mostly dyed deeper, implying that TPS++ learning has made text foreground playing a more important role in recognition. The results clearly demonstrate that the attention mechanism is appropriately formulated and the rectification can be well established at feature-level.

\subsection{Results and Comparisons}

We first assess how TPS++ is compatible with different backbones. Thus, three STR models adopting ResNet-31 \cite{yue2020robustscanner}, ResNet-45 \cite{ABInet21CVPR} and ViT \cite{dosovitskiy2020image} are selected. The first two can be directly used while for ViT a simple accommodation is required. The results are shown in Tab.\ref{tab:backbone}. Performance improvements are steadily observed no matter which backbone is employed. When looking into the irregular datasets, TPS++ receives performance gains of 1.7-5.5\%, 3.6-5.0\% and 2.4-4.7\% on IC15, SVTP and CT80 across the three backbones, respectively. The encouraging results indicate that TPS++ has a strong adaptation to backbone changes. It also proves the effectiveness of attention-based feature-level rectification. 

\begin{table}[t]
\resizebox{1.0\linewidth}{!}{
\begin{tabular}{cccccccc}
\hline
\multirow{2}{*}{Backbone}  & \multirow{2}{*}{TPS++}  & \multirow{2}{*}{IIIT5k} & \multirow{2}{*}{SVT}  & \multirow{2}{*}{IC13} & \multirow{2}{*}{IC15} & \multirow{2}{*}{SVTP}  & \multirow{2}{*}{CUTE} \\
\\
\hline
ViT\shortcite{dosovitskiy2020image} & & 89.8 & 85.2 & 89.3 & 69.6 & 75.8 & 78.1 \\
ViT\shortcite{dosovitskiy2020image} &\checkmark & 92.2 & 88.3 & 90.8 & 75.1 & 80.0 & 81.3 \\ 
ResNet-31\shortcite{yue2020robustscanner} &  & 94.1 & 90.9 & 92.4 & 77.4 & 81.2 & 84.4 \\ 
ResNet-31\shortcite{yue2020robustscanner} & \checkmark & 94.6 & 92.0 & 94.8 & 79.1 & 84.8 & 86.8 \\ 
ResNet-45\shortcite{ABInet21CVPR} & &91.5 &87.8 &90.1 &73.5 &78.6 & 80.6 \\ 
ResNet-45\shortcite{ABInet21CVPR} & \checkmark & 94.1 & 91.2 & 92.2 & 78.7 & 82.6 & 84.3 \\
\hline
\end{tabular}}
\caption{TPS++ evaluation on different backbones.}
\label{tab:backbone}
\end{table}

We then evaluate the performance of TPS++ with different recognizers. Concretely, CRNN \cite{ShiBY17crnn}, NRTR \cite{sheng2019nrtr} and ABINet \cite{ABInet21CVPR} are selected, which stand for CTC-based, attention-based and parallel decoding-based methods, respectively. Similarly, other components are kept unchanged. Tab.\ref{tab:recognizers} presents the results, where ABINet-LV is ABINet-V plus with the language model. Generally, TPS++ brings performance gains on all the recognizers especially in irregular datasets. The results also imply that TPS++ generalizes well across recognizers. Together with the results on backbones, it reliably verifies the plug-in property of TPS++.

\begin{table}[!htb]
\resizebox{1.0\linewidth}{!}{
\begin{tabular}{c|cc|cccccc}
\hline
       & TPS       & TPS++      & IIIT5K        & SVT           & IC13          & IC15          & SVTP          & CT80          \\ \hline
CRNN\shortcite{ShiBY17crnn}   &           &            & 92.1          & 87.3          & 91.8          & 77.5          & 77.4          & 83.4           \\
CRNN\shortcite{ShiBY17crnn}   & \checkmark         &            & 92.6          & 87.8          & 92.1 & 78.6          & 78.9          & 84.1           \\
CRNN\shortcite{ShiBY17crnn}   & \textbf{} & \textbf{\checkmark} & 92.9 & 88.8 & 92.8          & 80.3 & 80.6 & 84.4 \\ \hline
NRTR\shortcite{sheng2019nrtr}   &           &            & 96.1          & 94.4          & 96            & 84.7          & 87.8          & 88.9          \\
NRTR\shortcite{sheng2019nrtr}   & \checkmark         &            & 95.9          & 94.2          & 96.4          & 85.5          & 88.2          & 89.7          \\
NRTR\shortcite{sheng2019nrtr}   &           & \checkmark          & 96.3 & 94.6 & 96.6 & 85.7 & 89   & 92.4  \\ \hline
ABINet-LV\shortcite{ABInet21CVPR} &           &            & 96.2          & 93            & 97            & 85            & 88.5          & 89.2           \\
ABINet-LV\shortcite{ABInet21CVPR} & \checkmark         &            & 95.8          & 93.5          & 97.5          & 86            & 88.9          & 88.9         \\
ABINet-LV\shortcite{ABInet21CVPR} &           & \checkmark          & 96.3 & 94.3 & 97.8 & 86.5 & 89.6 & 89.6 \\ \hline
\end{tabular} }
\caption{TPS++ evaluation on different recognizers.}
\label{tab:recognizers}
\end{table}

In Tab.\ref{tab:sota}, we list the results of existing studies from the angle of both popular recognizers and rectifiers. When equipped with TPS++, both NRTR and ABINet-LV gain performance improvements and surpass almost all comparing methods. Specifically, 3 out of the 6 benchmarks obtained by ABINet-LV+TPS++ are new state of the art, while the cost is nearly 0.5M increase on parameters and 4.3ms on inference time compared to ABINet-LV. Both are marginal when compared to the differences among different models (Here both CRNN and NRTR are equipped with ResNet-45, a more powerful backbone compared to their raw implementation). When compared with ASTER, CRNN+TPS++ already gives nearly 3 percent accuracy improvements on irregular text, consuming just 74\% parameters and 23\% inference time. Meanwhile, ABINet-LV+TPS++ gives an even larger accuracy gap and still runs faster. The results clearly reveal that TPS++ is an accurate, efficient and universal tool towards better STR.

In Fig.\ref{fig:vis_point} we illustrate several successful and failed cases. The rectified image is also given by projecting the rectification back to the image. As seen, TPS++ gives a relative natural rectification even severe text distortions are presented, and thus gets the correct recognition. For the failure cases (the fourth line). They are mainly because the text is severely blurred. Despite being corrected to some extent, they are even unreadable for human beings, and recognizing such cases still remains a common difficulty for STR.

\section{Conclusion}
We have presented TPS++, an attention-enhanced rectifier for STR. It introduces the content-aware attention score to TPS formula, endowing the transformation with more flexibility. Moreover, it is a feature-level rectifier that partially shares the feature backbone with the recognizer, thus lightweight and fast. The experiments conducted on standard benchmarks basically validate our proposal, from which the effectiveness, efficiency and universality of TPS++ are well demonstrated. We hope that TPS++ will foster future research in STR, text spotting \cite{fang2022abinet++}, etc.

\section*{Acknowledgments}
The work was supported by National Key R\&D Program of China (No. 2020AAA0104500) and the National Natural Science Foundation of China (No. 62172103).

\bibliographystyle{named}
\bibliography{ijcai23}

\begin{thebibliography}{}

\bibitem[\protect\citeauthoryear{Baek \bgroup \em et al.\egroup
  }{2019}]{Baekwhats_wrong_19ICCV}
Jeonghun Baek, Geewook Kim, Junyeop Lee, Sungrae Park, Dongyoon Han, Sangdoo
  Yun, Seong~Joon Oh, and Hwalsuk Lee.
\newblock What is wrong with scene text recognition model comparisons? dataset
  and model analysis.
\newblock In {\em ICCV}, pages 4714--4722, 2019.

\bibitem[\protect\citeauthoryear{Bautista and
  Atienza}{2022}]{bautista2022parseq}
Darwin Bautista and Rowel Atienza.
\newblock Scene text recognition with permuted autoregressive sequence models.
\newblock In {\em ECCV}, pages 178--196, Cham, 10 2022. Springer Nature
  Switzerland.

\bibitem[\protect\citeauthoryear{Bhunia \bgroup \em et al.\egroup
  }{2021}]{bhunia2021jvsr}
Ayan~Kumar Bhunia, Aneeshan Sain, Amandeep Kumar, Shuvozit Ghose, Pinaki~Nath
  Chowdhury, and Yi-Zhe Song.
\newblock Joint visual semantic reasoning: Multi-stage decoder for text
  recognition.
\newblock In {\em ICCV}, pages 14920--14929, 2021.

\bibitem[\protect\citeauthoryear{Bookstein}{1989}]{bookstein1989tps}
Fred~L. Bookstein.
\newblock Principal warps: Thin-plate splines and the decomposition of
  deformations.
\newblock {\em IEEE Transactions on pattern analysis and machine intelligence},
  11(6):567--585, 1989.

\bibitem[\protect\citeauthoryear{Chen \bgroup \em et al.\egroup
  }{2021}]{Chen2021text}
Xiaoxue Chen, Lianwen Jin, Yuanzhi Zhu, Canjie Luo, and Tianwei Wang.
\newblock Text recognition in the wild: A survey.
\newblock {\em ACM Comput. Surv.}, 54(2), 2021.

\bibitem[\protect\citeauthoryear{Cheng \bgroup \em et al.\egroup
  }{2017}]{cheng2017focusing_attention}
Zhanzhan Cheng, Fan Bai, Yunlu Xu, Gang Zheng, Shiliang Pu, and Shuigeng Zhou.
\newblock Focusing attention: Towards accurate text recognition in natural
  images.
\newblock In {\em ICCV}, pages 5076--5084, 2017.

\bibitem[\protect\citeauthoryear{Dosovitskiy \bgroup \em et al.\egroup
  }{2020}]{dosovitskiy2020image}
Alexey Dosovitskiy, Lucas Beyer, Alexander Kolesnikov, Dirk Weissenborn,
  Xiaohua Zhai, Thomas Unterthiner, Mostafa Dehghani, Matthias Minderer, Georg
  Heigold, Sylvain Gelly, et~al.
\newblock An image is worth 16x16 words: Transformers for image recognition at
  scale.
\newblock {\em arXiv preprint arXiv:2010.11929}, 2020.

\bibitem[\protect\citeauthoryear{Du \bgroup \em et al.\egroup
  }{2022}]{du2022@svtr}
Yongkun Du, Zhineng Chen, Caiyan Jia, Xiaoting Yin, Tianlun Zheng, Chenxia Li,
  Yuning Du, and Yu{-}Gang Jiang.
\newblock {SVTR:} scene text recognition with a single visual model.
\newblock {\em IJCAI}, 2022.

\bibitem[\protect\citeauthoryear{Fang \bgroup \em et al.\egroup
  }{2018}]{fang2018attention}
Shancheng Fang, Hongtao Xie, Zheng-Jun Zha, Nannan Sun, Jianlong Tan, and
  Yongdong Zhang.
\newblock Attention and language ensemble for scene text recognition with
  convolutional sequence modeling.
\newblock In {\em ACM MM}, pages 248--256, 2018.

\bibitem[\protect\citeauthoryear{Fang \bgroup \em et al.\egroup
  }{2021}]{ABInet21CVPR}
Shancheng Fang, Hongtao Xie, Yuxin Wang, Zhendong Mao, and Yongdong Zhang.
\newblock Read like humans: Autonomous, bidirectional and iterative language
  modeling for scene text recognition.
\newblock In {\em CVPR}, pages 7094--7103, 2021.

\bibitem[\protect\citeauthoryear{Fang \bgroup \em et al.\egroup
  }{2022}]{fang2022abinet++}
Shancheng Fang, Zhendong Mao, Hongtao Xie, Yuxin Wang, Chenggang Yan, and
  Yongdong Zhang.
\newblock Abinet++: Autonomous, bidirectional and iterative language modeling
  for scene text spotting.
\newblock {\em IEEE Transactions on Pattern Analysis and Machine Intelligence},
  2022.

\bibitem[\protect\citeauthoryear{Gupta \bgroup \em et al.\egroup }{2016}]{ST}
Ankush Gupta, Andrea Vedaldi, and Andrew Zisserman.
\newblock Synthetic data for text localisation in natural images.
\newblock In {\em CVPR}, pages 2315--2324, 2016.

\bibitem[\protect\citeauthoryear{He \bgroup \em et al.\egroup
  }{2022}]{he2021S-GTR}
Yue He, Chen Chen, Jing Zhang, Juhua Liu, Fengxiang He, Chaoyue Wang, and
  Bo~Du.
\newblock Visual semantics allow for textual reasoning better in scene text
  recognition.
\newblock In {\em AAAI}, 2022.

\bibitem[\protect\citeauthoryear{Hu \bgroup \em et al.\egroup
  }{2020}]{hu2020gtc}
Wenyang Hu, Xiaocong Cai, Jun Hou, Shuai Yi, and Zhiping Lin.
\newblock Gtc: Guided training of ctc towards efficient and accurate scene text
  recognition.
\newblock In {\em AAAI}, volume~34, pages 11005--11012, 2020.

\bibitem[\protect\citeauthoryear{Jaderberg \bgroup \em et al.\egroup
  }{2014}]{MJor90K}
Max Jaderberg, Karen Simonyan, Andrea Vedaldi, and Andrew Zisserman.
\newblock Synthetic data and artificial neural networks for natural scene text
  recognition.
\newblock {\em arXiv preprint arXiv:1406.2227}, 2014.

\bibitem[\protect\citeauthoryear{Jaderberg \bgroup \em et al.\egroup
  }{2015}]{jaderberg2015stn}
Max Jaderberg, Karen Simonyan, Andrew Zisserman, et~al.
\newblock Spatial transformer networks.
\newblock In {\em NIPS}, volume~28, pages 2017--2025, 2015.

\bibitem[\protect\citeauthoryear{Karatzas \bgroup \em et al.\egroup
  }{2013}]{ICDAR2013}
Dimosthenis Karatzas, Faisal Shafait, Seiichi Uchida, Masakazu Iwamura,
  Lluis~Gomez i~Bigorda, Sergi~Robles Mestre, Joan Mas, David~Fernandez Mota,
  Jon~Almazan Almazan, and Lluis~Pere De~Las~Heras.
\newblock Icdar 2013 robust reading competition.
\newblock In {\em ICDAR}, pages 1484--1493, 2013.

\bibitem[\protect\citeauthoryear{Karatzas \bgroup \em et al.\egroup
  }{2015}]{ICDAR2015}
Dimosthenis Karatzas, Lluis Gomez-Bigorda, Anguelos Nicolaou, Suman Ghosh,
  Andrew Bagdanov, Masakazu Iwamura, Jiri Matas, Lukas Neumann,
  Vijay~Ramaseshan Chandrasekhar, Shijian Lu, et~al.
\newblock Icdar 2015 competition on robust reading.
\newblock In {\em ICDAR}, pages 1156--1160, 2015.

\bibitem[\protect\citeauthoryear{Lee and
  Osindero}{2016}]{lee2016attention_origin}
Chen-Yu Lee and Simon Osindero.
\newblock Recursive recurrent nets with attention modeling for ocr in the wild.
\newblock In {\em CVPR}, pages 2231--2239, 2016.

\bibitem[\protect\citeauthoryear{Li \bgroup \em et al.\egroup
  }{2017}]{li2017fcsem}
Yi~Li, Haozhi Qi, Jifeng Dai, Xiangyang Ji, and Yichen Wei.
\newblock Fully convolutional instance-aware semantic segmentation.
\newblock In {\em CVPR}, pages 2359--2367, 2017.

\bibitem[\protect\citeauthoryear{Li \bgroup \em et al.\egroup
  }{2019}]{li2019sar}
Hui Li, Peng Wang, Chunhua Shen, and Guyu Zhang.
\newblock Show, attend and read: A simple and strong baseline for irregular
  text recognition.
\newblock In {\em AAAI}, volume~33, pages 8610--8617, 2019.

\bibitem[\protect\citeauthoryear{Liao \bgroup \em et al.\egroup
  }{2019}]{liao2019two_dim_per}
Minghui Liao, Jian Zhang, Zhaoyi Wan, Fengming Xie, Jiajun Liang, Pengyuan Lyu,
  Cong Yao, and Xiang Bai.
\newblock Scene text recognition from two-dimensional perspective.
\newblock In {\em AAAI}, volume~33, pages 8714--8721, 2019.

\bibitem[\protect\citeauthoryear{Lin \bgroup \em et al.\egroup
  }{2021}]{lin2021stan}
Qingxiang Lin, Canjie Luo, Lianwen Jin, and Songxuan Lai.
\newblock Stan: A sequential transformation attention-based network for scene
  text recognition.
\newblock {\em Pattern Recognition}, 111:107692, 2021.

\bibitem[\protect\citeauthoryear{Liu \bgroup \em et al.\egroup
  }{2016}]{liu2016starnet}
Wei Liu, Chaofeng Chen, Kwan-Yee~K Wong, Zhizhong Su, and Junyu Han.
\newblock Star-net: a spatial attention residue network for scene text
  recognition.
\newblock In {\em BMVC}, volume~2, page~7, 2016.

\bibitem[\protect\citeauthoryear{Liu \bgroup \em et al.\egroup
  }{2018}]{liu2018charnet}
Wei Liu, Chaofeng Chen, and Kwan-Yee~K Wong.
\newblock Char-net: A character-aware neural network for distorted scene text
  recognition.
\newblock In {\em AAAI}, pages 7154--7161, 2018.

\bibitem[\protect\citeauthoryear{Long \bgroup \em et al.\egroup
  }{2021}]{long2021str_era}
Shangbang Long, Xin He, and Cong Yao.
\newblock Scene text detection and recognition: The deep learning era.
\newblock {\em International Journal of Computer Vision}, 129(1):161--184,
  2021.

\bibitem[\protect\citeauthoryear{Luo \bgroup \em et al.\egroup
  }{2019}]{cluo2019moran}
Canjie Luo, Lianwen Jin, and Zenghui Sun.
\newblock {MORAN}: A multi-object rectified attention network for scene text
  recognition.
\newblock {\em Pattern Recognition}, 90:109--118, 2019.

\bibitem[\protect\citeauthoryear{Mishra \bgroup \em et al.\egroup
  }{2012}]{IIIT5K}
Anand Mishra, Karteek Alahari, and CV~Jawahar.
\newblock Scene text recognition using higher order language priors.
\newblock In {\em BMVC}, pages 1--11, 2012.

\bibitem[\protect\citeauthoryear{Phan \bgroup \em et al.\egroup }{2013}]{SVT-P}
Trung~Quy Phan, Palaiahnakote Shivakumara, Shangxuan Tian, and Chew~Lim Tan.
\newblock Recognizing text with perspective distortion in natural scenes.
\newblock In {\em ICCV}, pages 569--576, 2013.

\bibitem[\protect\citeauthoryear{Qian \bgroup \em et al.\egroup
  }{2021}]{qian2021adaptive}
Ye~Qian, Long Chen, and Feng Su.
\newblock An adaptive rectification model for arbitrary-shaped scene text
  recognition.
\newblock page~91, 2021.

\bibitem[\protect\citeauthoryear{Risnumawan \bgroup \em et al.\egroup
  }{2014}]{CUTE80}
Anhar Risnumawan, Palaiahankote Shivakumara, Chee~Seng Chan, and Chew~Lim Tan.
\newblock A robust arbitrary text detection system for natural scene images.
\newblock {\em ESA}, 41(18):8027--8048, 2014.

\bibitem[\protect\citeauthoryear{Sheng \bgroup \em et al.\egroup
  }{2019}]{sheng2019nrtr}
Fenfen Sheng, Zhineng Chen, and Bo~Xu.
\newblock Nrtr: A no-recurrence sequence-to-sequence model for scene text
  recognition.
\newblock In {\em ICDAR}, pages 781--786, 2019.

\bibitem[\protect\citeauthoryear{Shi \bgroup \em et al.\egroup
  }{2016}]{shi2016robust_auto_rect}
Baoguang Shi, Xinggang Wang, Pengyuan Lyu, Cong Yao, and Xiang Bai.
\newblock Robust scene text recognition with automatic rectification.
\newblock In {\em CVPR}, pages 4168--4176, 2016.

\bibitem[\protect\citeauthoryear{Shi \bgroup \em et al.\egroup
  }{2017}]{ShiBY17crnn}
Baoguang Shi, Xiang Bai, and Cong Yao.
\newblock An end-to-end trainable neural network for image-based sequence
  recognition and its application to scene text recognition.
\newblock {\em IEEE Transactions on Pattern Analysis and Machine Intelligence},
  39(11):2298--2304, 2017.

\bibitem[\protect\citeauthoryear{Shi \bgroup \em et al.\egroup
  }{2018}]{shi2018aster}
Baoguang Shi, Mingkun Yang, Xinggang Wang, Pengyuan Lyu, Cong Yao, and Xiang
  Bai.
\newblock Aster: An attentional scene text recognizer with flexible
  rectification.
\newblock {\em IEEE Transactions on Pattern Analysis and Machine Intelligence},
  41(9):2035--2048, 2018.

\bibitem[\protect\citeauthoryear{Wang \bgroup \em et al.\egroup }{2011}]{SVT}
Kai Wang, Boris Babenko, and Serge Belongie.
\newblock End-to-end scene text recognition.
\newblock In {\em ICCV}, pages 1457--1464, 2011.

\bibitem[\protect\citeauthoryear{Wang \bgroup \em et al.\egroup
  }{2021}]{wang2021FTO}
Yuxin Wang, Hongtao Xie, Shancheng Fang, Jing Wang, Shenggao Zhu, and Yongdong
  Zhang.
\newblock From two to one: A new scene text recognizer with visual language
  modeling network.
\newblock In {\em ICCV}, pages 14174--14183, 2021.

\bibitem[\protect\citeauthoryear{Wang \bgroup \em et al.\egroup
  }{2022}]{wang2022petr}
Yuxin Wang, Hongtao Xie, Shancheng Fang, Mengting Xing, Jing Wang, Shenggao
  Zhu, and Yongdong Zhang.
\newblock Petr: Rethinking the capability of transformer-based language model
  in scene text recognition.
\newblock {\em IEEE Transactions on Image Processing}, 31:5585--5598, 2022.

\bibitem[\protect\citeauthoryear{Woo \bgroup \em et al.\egroup
  }{2018}]{woo2018cbam}
Sanghyun Woo, Jongchan Park, Joon-Young Lee, and In~So Kweon.
\newblock Cbam: Convolutional block attention module.
\newblock In {\em ECCV}, pages 3--19, 2018.

\bibitem[\protect\citeauthoryear{Xing \bgroup \em et al.\egroup
  }{2019}]{xing2019convolutional}
Linjie Xing, Zhi Tian, Weilin Huang, and Matthew~R Scott.
\newblock Convolutional character networks.
\newblock In {\em ICCV}, pages 9126--9136, 2019.

\bibitem[\protect\citeauthoryear{Yan \bgroup \em et al.\egroup
  }{2021}]{yan2021primitive}
Ruijie Yan, Liangrui Peng, Shanyu Xiao, and Gang Yao.
\newblock Primitive representation learning for scene text recognition.
\newblock In {\em CVPR}, pages 284--293, 2021.

\bibitem[\protect\citeauthoryear{Yang \bgroup \em et al.\egroup
  }{2019}]{yang2019symmetry}
Mingkun Yang, Yushuo Guan, Minghui Liao, Xin He, Kaigui Bian, Song Bai, Cong
  Yao, and Xiang Bai.
\newblock Symmetry-constrained rectification network for scene text
  recognition.
\newblock In {\em ICCV}, pages 9147--9156, 2019.

\bibitem[\protect\citeauthoryear{Yang \bgroup \em et al.\egroup
  }{2022}]{yang2022reading}
Mingkun Yang, Minghui Liao, Pu~Lu, Jing Wang, Shenggao Zhu, Hualin Luo,
  Qi~Tian, and Xiang Bai.
\newblock Reading and writing: Discriminative and generative modeling for
  self-supervised text recognition.
\newblock In {\em ACM MM}, pages 4214--4223, 2022.

\bibitem[\protect\citeauthoryear{Yu \bgroup \em et al.\egroup
  }{2020}]{SRNyu2020towards}
Deli Yu, Xuan Li, Chengquan Zhang, Tao Liu, Junyu Han, Jingtuo Liu, and Errui
  Ding.
\newblock Towards accurate scene text recognition with semantic reasoning
  networks.
\newblock In {\em CVPR}, pages 12113--12122, 2020.

\bibitem[\protect\citeauthoryear{Yu \bgroup \em et al.\egroup
  }{2023}]{yu2023structextv2}
Yuechen Yu, Yulin Li, Chengquan Zhang, Xiaoqiang Zhang, Zengyuan Guo, Xiameng
  Qin, Kun Yao, Junyu Han, Errui Ding, and Jingdong Wang.
\newblock Structextv2: Masked visual-textual prediction for document image
  pre-training.
\newblock {\em arXiv preprint arXiv:2303.00289}, 2023.

\bibitem[\protect\citeauthoryear{Yue \bgroup \em et al.\egroup
  }{2020}]{yue2020robustscanner}
Xiaoyu Yue, Zhanghui Kuang, Chenhao Lin, Hongbin Sun, and Wayne Zhang.
\newblock Robustscanner: Dynamically enhancing positional clues for robust text
  recognition.
\newblock In {\em ECCV}, pages 135--151, 2020.

\bibitem[\protect\citeauthoryear{Zhan and Lu}{2019}]{zhan2019esir}
Fangneng Zhan and Shijian Lu.
\newblock Esir: End-to-end scene text recognition via iterative image
  rectification.
\newblock In {\em CVPR}, pages 2059--2068, 2019.

\bibitem[\protect\citeauthoryear{Zhang \bgroup \em et al.\egroup
  }{2021}]{zhang2021spin}
Chengwei Zhang, Yunlu Xu, Zhanzhan Cheng, Shiliang Pu, Yi~Niu, Fei Wu, and
  Futai Zou.
\newblock Spin: Structure-preserving inner offset network for scene text
  recognition.
\newblock In {\em AAAI}, volume~35, pages 3305--3314, 2021.

\bibitem[\protect\citeauthoryear{Zheng \bgroup \em et al.\egroup
  }{2021}]{zheng2021cdistnet}
Tianlun Zheng, Zhineng Chen, Shancheng Fang, Hongtao Xie, and Yu-Gang Jiang.
\newblock Cdistnet: Perceiving multi-domain character distance for robust text
  recognition.
\newblock {\em arXiv preprint arXiv:2111.11011}, 2021.

\bibitem[\protect\citeauthoryear{Zhong \bgroup \em et al.\egroup
  }{2022}]{zhong2022sgbanet}
Dajian Zhong, Shujing Lyu, Palaiahnakote Shivakumara, Bing Yin, Jiajia Wu,
  Umapada Pal, and Yue Lu.
\newblock Sgbanet: Semantic gan and balanced attention network for arbitrarily
  oriented scene text recognition.
\newblock {\em ECCV}, 2022.

\end{thebibliography}
\end{document}